\documentclass[12pt, anon]{l4dc2024}

\newcommand{\R}[1]{\mathbf{R}^{#1}}

\SetCommentSty{mycommfont}

\usepackage{optidef}
\usepackage{wrapfig}
\usepackage{float}
\usepackage{hyperref}

\usepackage{sidecap, caption}
\sidecaptionvpos{figure}{t}

\usepackage[utf8]{inputenc}
\usepackage{pgfplots}
\usepackage{subcaption}

\DeclareUnicodeCharacter{2212}{−}
\usepgfplotslibrary{groupplots,dateplot}
\usetikzlibrary{patterns,shapes.arrows}
\pgfplotsset{compat=newest}

\title[Efficient Online Learning of Contact Force Models]{Efficient Online Learning of Contact Force Models \\ for Connector Insertion}
\usepackage{times}

\author{%
 \Name{Kevin Tracy} \Email{ktracy@cmu.edu}\\
 \Name{Zachary Manchester} \Email{zacm@cmu.edu}\\
 \addr The Robotics Institute, Carnegie Mellon University
 \AND
 \Name{Ajinkya Jain} \Email{ajinkyajain@intrinsic.ai}\\
 \Name{Keegan Go} \Email{keegango@intrinsic.ai} \\
 \Name{Stefan Schaal} \Email{sschaal@intrinsic.ai} \\
\addr [Google] Intrinsic
 \AND 
 \Name{Tom Erez} \Email{etom@google.com} \\ 
 \Name{Yuval Tassa} \Email{tassa@google.com} \\
 \addr Google DeepMind Robotics
}

\begin{document}

\maketitle

\begin{abstract}%
Contact-rich manipulation tasks with stiff frictional elements like connector insertion are difficult to model with rigid-body simulators.  In this work, we propose a new approach for modeling these environments by learning a quasi-static contact force model instead of a full simulator. Using a feature vector that contains information about the configuration and control, we find a linear mapping adequately captures the relationship between this feature vector and the sensed contact forces.  A novel Linear Model Learning (LML) algorithm is used to solve for the globally optimal mapping in real time without any matrix inversions, resulting in an algorithm that runs in nearly constant time on a GPU as the model size increases. 
We validate the proposed approach for connector insertion both in simulation and hardware experiments, where the learned model is combined with an optimization-based controller to achieve smooth insertions in the presence of misalignments and uncertainty. Our website featuring videos, code, and more materials is available at \url{https://model-based-plugging.github.io/}.
\end{abstract}
%
\section{Introduction} \label{sect:intro}
%
Model-based control for robotic systems is incredibly effective when the model of the system is accurate \citep{tedrake2014}. In many instances, the model can be made accurate through system identification, where experimental data is used with a parametrized physics model to estimate geometric, mass, contact, and friction properties \citep{galrinho2016, hoburg2009}.  In scenarios where contact-free rigid-body dynamics accurately capture the behavior of the robot, system identification can enable highly performant model-based control. For robots in contact-rich settings where they make and break contact with the environment, system identification and simulation become much more challenging due to discontinuities and inevitable approximations in the physics modeling \citep{zhao2020}.

Modern robotics simulators like MuJoCo \citep{todorov2012}, Bullet \citep{coumans2015}, Dart \citep{lee2018}, Dojo \citep{howell2022}, and Isaac Gym \citep{makoviychuk2021}, approximate all bodies as rigid, where friction is either the idealized Coulomb friction or a closely related approximation. This results in a significant gap between simulation and reality, even with the use of system identification \citep{acary2018, horak2019, chatterjee, elandt2019}. Accurate simulation of contact-rich manipulation tasks is challenging due to deformability in the bodies, approximate geometries, and complex friction behavior not captured by Coulomb friction \citep{drumwright2010}.
\begin{figure}[t]
\centering 
\begin{tikzpicture}
    \draw (0, 0) node[inner sep=0] {\includegraphics[width=11cm]{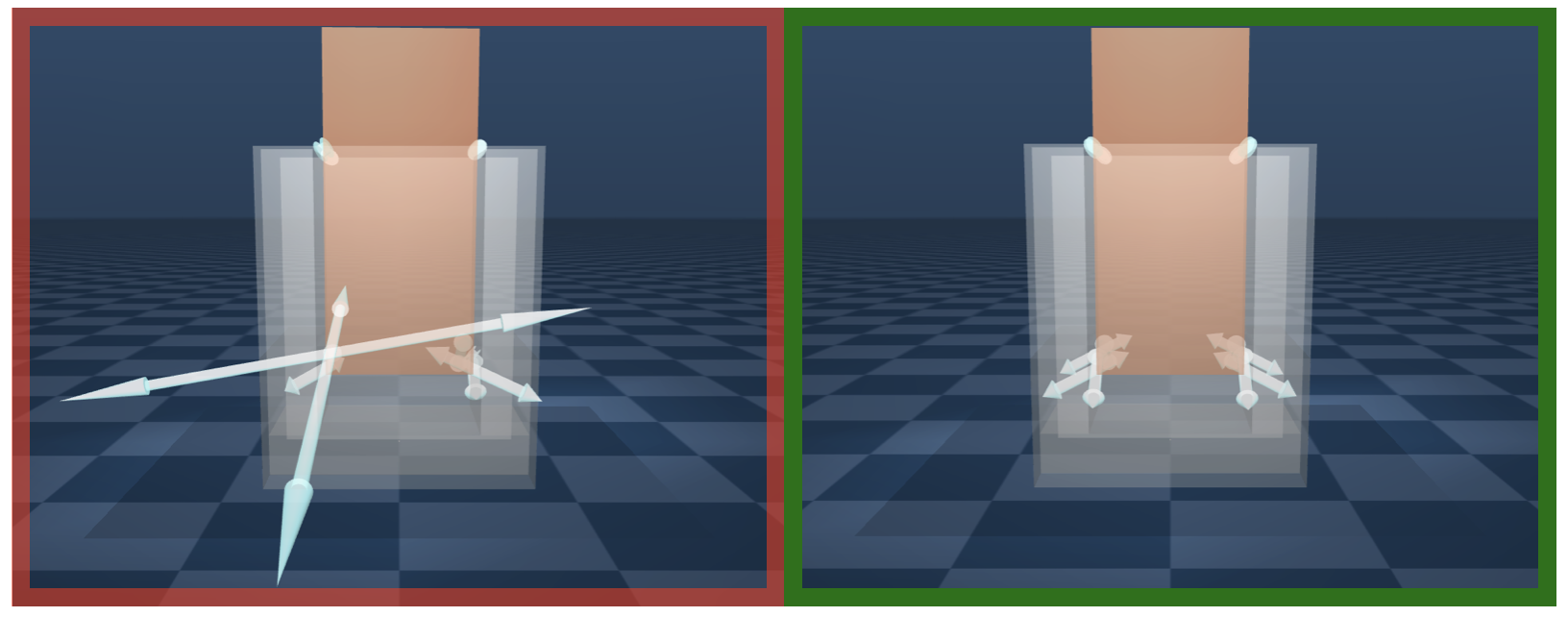}};
    \node at (-1.0,-1.6) [fill=red!20, text=black] {Baseline};
    \node at (4.7,-1.6) [fill=green!20, text=black] {Ours};
\end{tikzpicture}
\caption{Inserting a significantly misaligned connector in MuJoCo, where a scripted insertion (left) results in large contact forces shown in white. During a short calibration sequence, a model is learned that predicts these contact forces as a function of the configuration of the connector and applied control wrench. A model-based controller is then able to solve for actions that minimize these forces and guide the connector to a smooth insertion (right). Our approach is fast, data efficient, robust to misalignments and uncertainty, and does not require any a priori knowledge of the plug.
}
\label{fig:mujoco}
\end{figure}

This paper focuses on the problem of robotic connector insertion, which displays rich contact interactions between a plug and its corresponding socket. This task is difficult to simulate accurately because of the precision required (being off by a millimeter results in jamming), and the contact/friction interactions between the two slightly deformable objects. The point-contact assumption that rigid-body simulators make begins to break down as the contact area dynamically changes.  As a result, accurately simulating the insertion is challenging, and a policy that relies on the accuracy of this simulation is poorly suited for the diverse range of connectors available. Furthermore, even if a single connector style is modeled in great detail, there exists a significant amount of variability amongst connectors of the exact same specification.

The difficulty of the connector-insertion problem has motivated multiple approaches, some model-based and others model-free. The authors in \cite{zhao2022} propose an offline model-free meta-reinforcement-learning algorithm to pre-train policies for insertion tasks using offline demonstration data and then adapt them to new tasks using online fine tuning. In \cite{fu2023}, the insertion is divided into an alignment phase and an insertion phase. During the alignment phase, a policy that relies on tactile feedback is used to line up the connector, and insertion is achieved with a separate RGB-image-based policy. In \cite{nair2023}, a simulation is used with domain randomization to train a robust policy over a wide variety of connectors, enabling generalization; and in \cite{tang2023} a simulation-aware policy training style is used to prioritize experiences that align with the physics model. Alternatively, in \cite{kang2022}, a policy is used solely to generate a search pattern that systematically inserts connectors in the presence of model and state uncertainty. In all of these approaches, significant time and effort must be devoted to offline training from experiments, making rapid adaptation to new plugs challenging.

In this work, we present a different approach to model-based connector insertion that relies on a learned quasi-static \textit{contact-force model} instead of a full simulator.  Many contact-rich robotic manipulation tasks can be assumed to be quasi-static, meaning the system moves slowly enough that the Coriolis forces and accelerations in the dynamics are small enough to be ignored \citep{mason2001a, halm2019}. This approximation allows us to describe the state of the robot solely by its configuration (without velocity).  When compared to full second-order dynamics simulation, quasi-static simulators are more stable, simpler, and have shown great promise for use in contact-rich manipulation planning \citep{pang2023,suh2020}.

When analyzing experimental data, a linear model with an engineered feature vector was found to be expressive enough to accurately predict contact forces on the connector during an insertion task. The linearity (in the features) of this model allows for an entirely decoupled model-learning process, where a novel Linear Model Learning (LML) filter is developed for learning the maximum a posteriori estimate to global optimality in a recursive fashion. The resulting algorithm is entirely free of matrix inversions, and the run time is nearly constant on a GPU as the size of the model increases. This model is then used in a convex-optimization-based control policy where the connector is smoothly inserted. The contributions of this paper include:
\begin{itemize}
\itemsep0em 
    \item A validated contact-force model that linearly maps a feature vector with information about the configuration and control to predicted force torque sensor values during a connector insertion.
    \item The Linear Model Learning (LML) algorithm that recursively updates the optimal estimated contact force model without any matrix inversions. 
    \item A model-based controller that uses this contact force model to achieve smooth connector insertions.
\end{itemize}
%
\section{Feature-Based Contact Force Model} \label{sect:force-model}
%
Conventionally, physics simulations used in model-based control take the form of a discrete dynamics function $x_{k+1} = f(x_k, u_k)$, where the current state, $x_k$, and control, $u_k$, are used to calculate the state at the next time step, $x_{k+1}$. The state of the system, $x$, is normally comprised of a configuration, $q \in \R{n_q}$, and velocity, $v \in \R{n_v}$ \citep{tedrake2014}. For systems modeled with standard rigid-body dynamics, the full second-order dynamics depend on both the configuration of the system as well as the velocity. However, for the damped,  slow-moving systems common in contact-rich manipulation, velocity-dependent terms like Coriolis forces can be small enough to be effectively ignored \citep{mason2001a}. The quasi-static assumption also simplifies the actuator dynamics of impedance-controlled robots, where the behavior of the controller is directly incorporated as part of the model \citep{pang2018}. 

For quasi-static contact-rich tasks where the end effector of the robot is in constant contact with the environment, we can learn a contact-force model and still maintain all of the predictive power of a full simulator. To accurately represent the dynamics in this regime, all we need is a model that takes in the current configuration of the robot and applied control, and outputs the predicted force-torque sensor reading.  With the ability to predict how the force-torque sensor responds to control inputs, a model-based controller is able to fully reason about the contact environment. 

The model we seek to learn takes the configuration of the system, the control, $u \in \R{n_u}$, and the contact force torque measurement, $\tilde{y} \in \R{n_y}$ as inputs. Specifically, we are going to map the configuration and control into a feature vector, $w \in \R{n_w}$, using an arbitrary mapping function,
\begin{align}
    w = \Omega(q, u), \label{eq:feature_vector_mapping}
\end{align}
after which we learn a \textit{linear} mapping that transforms the feature vector into the predicted force, $f \in \R{3}$, and torque, $\tau \in \R{3}$, on the connector, $\tilde{y} = [f^T, \tau^T]^T \in \R{6}$:
\begin{align}
    \tilde{y} \approx \tilde{G}w \label{eq:linmod},
\end{align}
where $\tilde{G} \in \R{n_y \times n_w}$. There are multiple benefits to representing our force model as a linear function of the feature vector, both in terms of system identification and control. For system identification, the resulting optimization problem is convex and can be solved to global optimality in real time with minimal computational cost. For control, as long as the feature vector in \eqref{eq:feature_vector_mapping} is linear in $u$, the predicted force and torque can be minimized with convex optimization. This is true even in the presence of a feature vector that is highly nonlinear in $q$.
%
\section{Linear Model Learning} \label{sect:model-learning}
%
Given data from a trajectory, a maximum-likelihood estimate (MLE) of the linear mapping $\tilde{G}$ can be computed with numerical optimization by penalizing differences between the predicted and observed sensor measurements \citep{xinjilefu2014}. To express this, we consider our contact force model from \eqref{eq:linmod} with a force-torque sensor noise covariance $R \in \mathbf{S}_+^{n_y}$, and introduce the following optimization problem:
\begin{mini}
{\tilde{G}}{\sum_{k=1}^{m} \|\tilde{G}w_k - \tilde{y}_k\|_{R^{-1}}^2, }{\label{full_opt}}{}
\end{mini}
where the subscript $k$ denotes the quantity from the $k^{\text{th}}$ timestep, $m$ is the length of the trajectory, and $\|\nu \|_{R^{-1}}^2 = \nu^TR^{-1}\nu$. It is worth pointing out that the primal variable in \eqref{full_opt} is a matrix and not a vector, so it is not a canonical least squares problem. That being said, the linearity of the model still makes this an unconstrained convex optimization problem, where a closed-form solution exists \citep{boyd2004}. However, for systems where the dimension of the feature vector is large, this problem can be both expensive to solve and numerically ill-conditioned when the trajectory is long \citep{benesty2005}.

There is a long history of solving variants of \eqref{full_opt} from data, starting with the Ho-Kalman algorithm \citep{ho1966} to more modern versions as shown in \cite{galrinho2016, ghahramani,hazan2017,hardt2019}. Instead of solving this problem directly, the following subsections walk through a method for decomposing \eqref{full_opt} into a set of smaller decoupled problems that can be solved in parallel, as well as a recursive method for solving this problem online in a scalable and numerically robust way that completely avoids matrix inversions.
\subsection{Decoupled Model Learning}
The first step in decoupling \eqref{full_opt} into a set of $n_y$ smaller problems is to take a Cholesky decomposition of the inverse sensor covariance, $R^{-1}$, such that $R^{-1} = LL^T$, where $L \in \R{n_y \times n_y}$ is the lower-triangular Cholesky factor. From here, \eqref{full_opt} can be re-written as the following:
\begin{mini}
{\tilde{G}}{\sum_{k=1}^{m} \|L^T \tilde{G}w_k - L^T\tilde{y}_k\|_2^2  }{\label{full_opt_ls}}{}
\end{mini}
Next, we introduce new variables $G = L^T \tilde{G}$ and $y = L^T \tilde{y}$ that decouple each row of $G$ and $y$ from the others:
\begin{mini}
{{G}}{\sum_{k=1}^{m} \|Gw_{k} - y_{k}\|_2^2,  }{\label{full_opt_ls_2}}{}
\end{mini}
where the solution to \eqref{full_opt_ls} can be recovered from the solution to \eqref{full_opt_ls_2} by simply transforming our matrix back with $\tilde{G} = L^{-T}G$. The inverse $L^{-T}$ is computed once offline for a specific sensor and stored. 

Next, we add regularization to the problem to improve conditioning and ensure that values in $G$ are of reasonable magnitudes. To do this, a quadratic regularizer is added in the following manner:
\begin{mini}
{{G}}{\sum_{k=1}^{m} \|Gw_{k} - y_{k}\|_2^2 + \|G b\|_2^2, }{\label{full_opt_ls_3}}{}
\end{mini}
where $b \in \R{n_w}$ contains the regularization weights. This allows for \eqref{full_opt_ls_3} to be decoupled into a set of $n_y$ problems where each row of $G$ is solved for entirely independently.  The resulting optimization problem over the $j^{\text{th}}$ row of $G$, and $y$ can now be formulated:
\begin{mini}
{g^{(j)}}{\sum_{k=1}^{m} \|(g^{(j)})^Tw_k - y_k^{(j)}\|_2^2 +  (b^T g^{(j)})^2, }{\label{mini_opt}}{}
\end{mini}
where the $g^{(j)}$ corresponds to the $j^{\text{th}}$ row of $G$.  This allows us to solve $n_y$ parallel instances of \eqref{mini_opt} with $n_w$ decision variables each, instead of one large problem with $n_y \cdot n_w$ variables.  It is important to note that there are no approximations made between \eqref{full_opt_ls_3} and \eqref{mini_opt}, and that they will always recover the same solution.
%
\section{Linear Model Learning with a Kalman Filter}
%
The problem posed in \eqref{mini_opt} is an unconstrained convex optimization problem whose solution can be computed by solving a single linear system, with a solution complexity that is cubic in the length of the feature vector, $n_w$. In scenarios where online, real-time system identification is desired, a recursive method can be used that updates the solution for each new sensor measurement. One option for solving \eqref{mini_opt} in a recursive fashion would be to use a standard Recursive Least Squares (RLS) algorithm. While this method can work, it has three major downsides: Long trajectories present numerical instability, adaptively updating the regularizer is expensive and approximate, and most importantly, there is no way to directly reason about process noise on the parameters \citep{liavas1998,liavas1999,benesty2005}. 

To tackle these issues, we formulate our model-learning problem as a dynamical system and use a Kalman Filter to estimate each row of $G$ independently. First, the connection between the canonical Kalman Filter and its corresponding optimization problem must be made \citep{kalman1960}. 
For a canonical dynamical system with a state $x\in\R{n_x}$ and measurement $z \in \R{n_z}$, the dynamics and measurement functions are described by,
\begin{align}
    x_{k+1} &= Ax_k + q_k, & q &\sim \mathcal{N}(0, V), \label{vanilla_kf_1}\\ 
    z_{k} &= Cx_{k} + r_k, & r&\sim \mathcal{N}(0, W),\label{vanilla_kf_2}
\end{align}
where $q$ and $r$ are the (unknown) process and sensor noises, and $x$ is the (unknown) state being estimated. The Kalman Filter is then initialized with a Gaussian belief over the initial condition, $x_0 \sim \mathcal{N}(\mu_{ic}, \Sigma_{ic})$. Since the initial belief, the process noise, and the sensor noise are all assumed to be Gaussian, the maximum a posteriori (MAP) problem takes the form of an unconstrained convex quadratic program:
\begin{mini}
{\mu_{0:m}}{ \|\mu_{ic} - \mu_0\|_{\Sigma_{ic}^{-1}}^2 + \sum_{k=0}^{m-1}  \|\mu_{k+1} - A \mu_{k}\|_{Q^{-1}}^2 + \|z_{k+1} - C \mu_{k+1}\|_{R^{-1}}^2, }{\label{map}}{}
\end{mini}
where the trajectory $\mu_{0:m}$ that minimizes the covariance-weighted errors in the dynamics, the sensor, and the initial belief is solved for.  This is the optimization problem that the Kalman Filter solves (to global optimality) in a recursive fashion. 
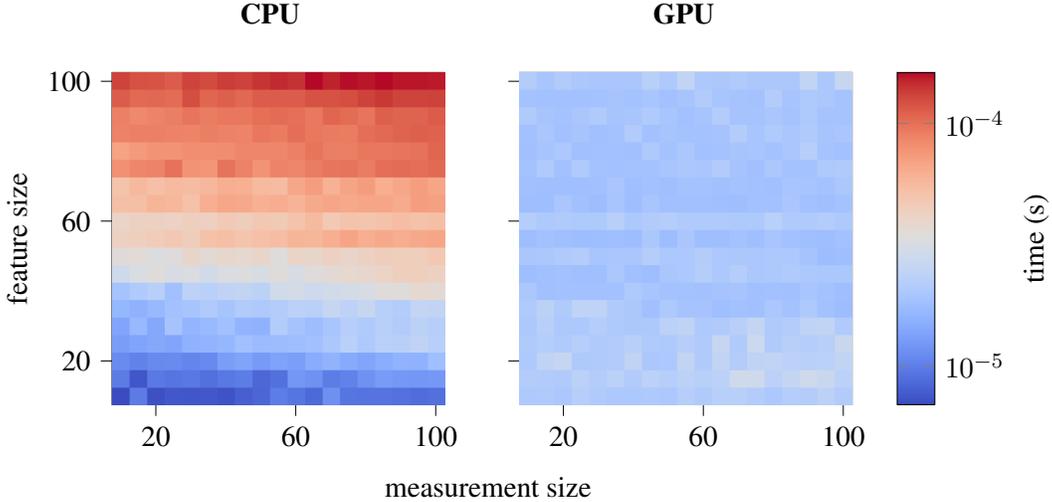
\begin{figure}[t]
    \centering
\begin{tikzpicture}
\node[] at (5,-1.1) {measurement size};
\node[] at (2.1,5.2) {\textbf{CPU}};
\node[] at (7.6,5.2) {\textbf{GPU}};

\definecolor{darkgray176}{RGB}{176,176,176}

\begin{groupplot}[group style={group size=2 by 1}]
\nextgroupplot[
tick align=outside,
tick pos=left,
x grid style={darkgray176},
xmin=-0.5, xmax=18.5,
xtick style={color=black},
xtick={2,10,18},
xticklabels={20,60,100},
y dir=reverse,
y grid style={darkgray176},
ylabel={feature size},
ymin=-0.5, ymax=18.5,
ytick style={color=black},
ytick={0,8,16},
yticklabels={100,60,20},
width=6cm,
height=6cm
]
\addplot graphics [includegraphics cmd=\pgfimage,xmin=-0.5, xmax=18.5, ymin=18.5, ymax=-0.5] {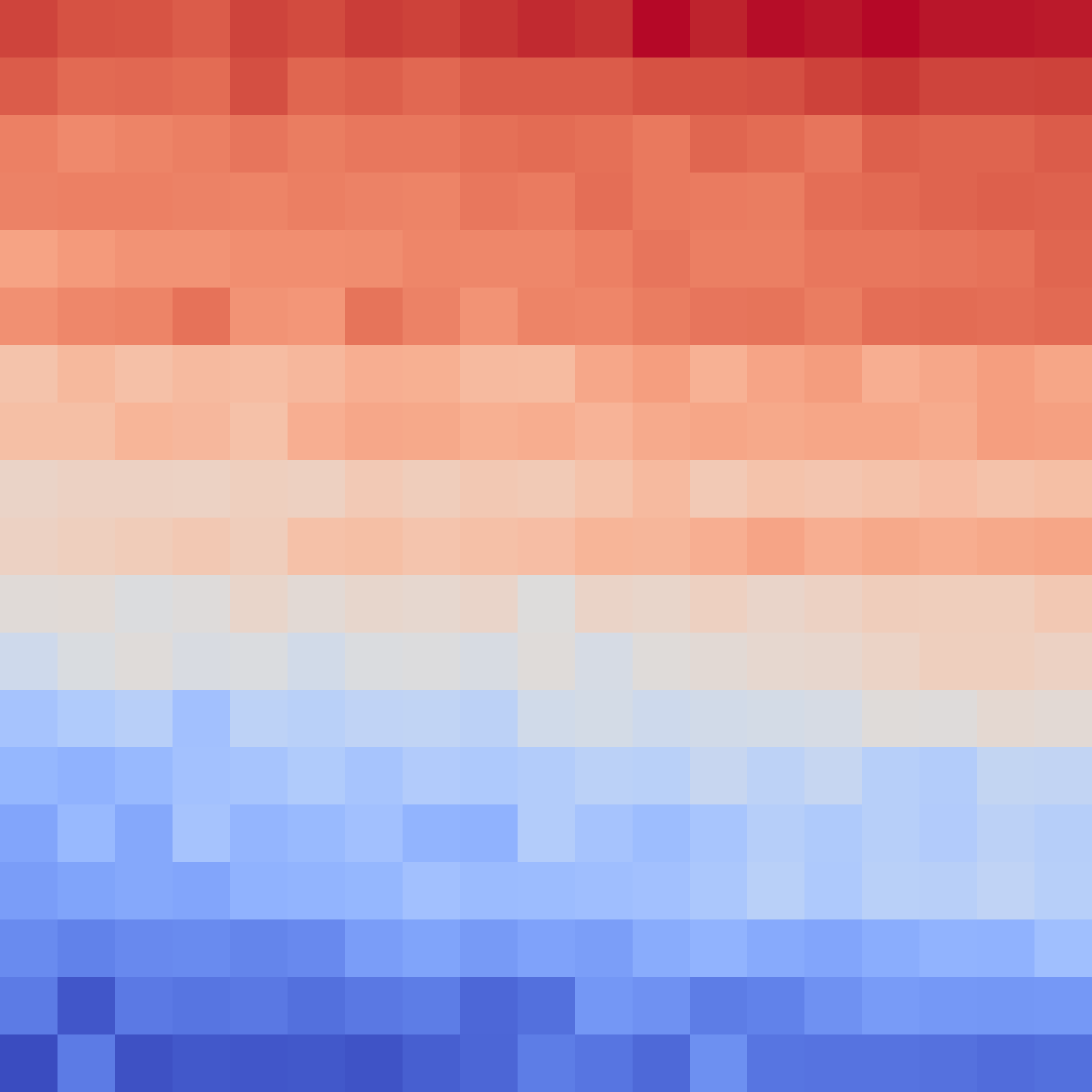};

\nextgroupplot[
scaled y ticks=manual:{}{\pgfmathparse{#1}},
colorbar,
colorbar style={ylabel=log(time (s)), ylabel style = {yshift=0pt},ytick={-5,-4},yticklabels={$10^{-5}$,$10^{-4}$},ylabel={time (s)}},
colormap={mymap}{[1pt]
  rgb(0pt)=(0.2298057,0.298717966,0.753683153);
  rgb(1pt)=(0.26623388,0.353094838,0.801466763);
  rgb(2pt)=(0.30386891,0.406535296,0.84495867);
  rgb(3pt)=(0.342804478,0.458757618,0.883725899);
  rgb(4pt)=(0.38301334,0.50941904,0.917387822);
  rgb(5pt)=(0.424369608,0.558148092,0.945619588);
  rgb(6pt)=(0.46666708,0.604562568,0.968154911);
  rgb(7pt)=(0.509635204,0.648280772,0.98478814);
  rgb(8pt)=(0.552953156,0.688929332,0.995375608);
  rgb(9pt)=(0.596262162,0.726149107,0.999836203);
  rgb(10pt)=(0.639176211,0.759599947,0.998151185);
  rgb(11pt)=(0.681291281,0.788964712,0.990363227);
  rgb(12pt)=(0.722193294,0.813952739,0.976574709);
  rgb(13pt)=(0.761464949,0.834302879,0.956945269);
  rgb(14pt)=(0.798691636,0.849786142,0.931688648);
  rgb(15pt)=(0.833466556,0.860207984,0.901068838);
  rgb(16pt)=(0.865395197,0.86541021,0.865395561);
  rgb(17pt)=(0.897787179,0.848937047,0.820880546);
  rgb(18pt)=(0.924127593,0.827384882,0.774508472);
  rgb(19pt)=(0.944468518,0.800927443,0.726736146);
  rgb(20pt)=(0.958852946,0.769767752,0.678007945);
  rgb(21pt)=(0.96732803,0.734132809,0.628751763);
  rgb(22pt)=(0.969954137,0.694266682,0.579375448);
  rgb(23pt)=(0.966811177,0.650421156,0.530263762);
  rgb(24pt)=(0.958003065,0.602842431,0.481775914);
  rgb(25pt)=(0.943660866,0.551750968,0.434243684);
  rgb(26pt)=(0.923944917,0.49730856,0.387970225);
  rgb(27pt)=(0.89904617,0.439559467,0.343229596);
  rgb(28pt)=(0.869186849,0.378313092,0.300267182);
  rgb(29pt)=(0.834620542,0.312874446,0.259301199);
  rgb(30pt)=(0.795631745,0.24128379,0.220525627);
  rgb(31pt)=(0.752534934,0.157246067,0.184115123);
  rgb(32pt)=(0.705673158,0.01555616,0.150232812)
},
point meta max=-3.79,
point meta min=-5.16,
tick align=outside,
tick pos=left,
x grid style={darkgray176},
xmin=-0.5, xmax=18.5,
xtick style={color=black},
xtick={2,10,18},
xticklabels={20,60,100},
y dir=reverse,
y grid style={darkgray176},
ymin=-0.5, ymax=18.5,
ytick style={color=black},
ytick={0,8,16},
yticklabels={,,},
width=6cm,
height=6cm,
]
\addplot graphics [includegraphics cmd=\pgfimage,xmin=-0.5, xmax=18.5, ymin=18.5, ymax=-0.5] {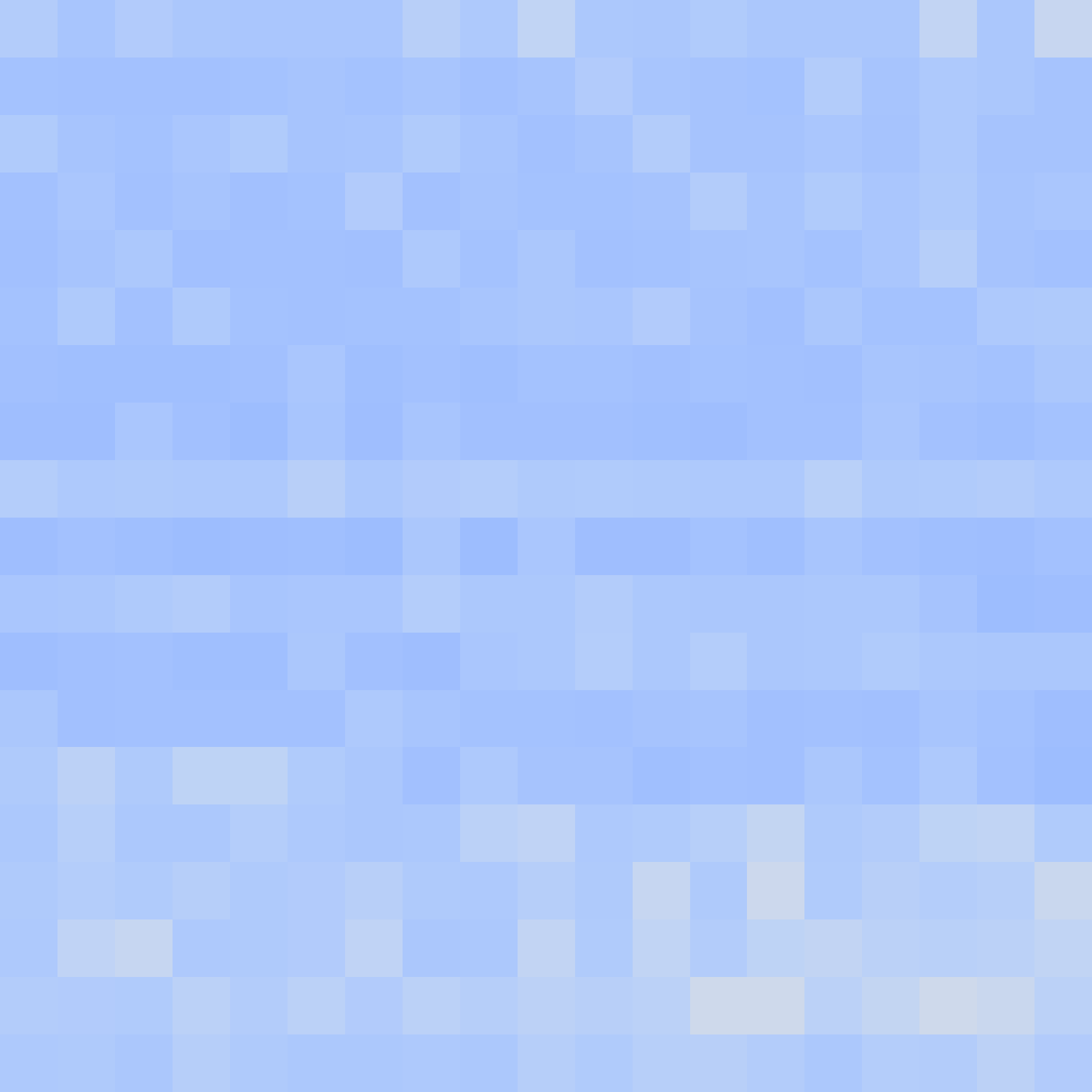};
\end{groupplot}

\end{tikzpicture}
    \caption{A comparison between run times of the LML algorithm on a CPU and an NVIDIA V100 GPU. The LML algorithm is comprised entirely of highly parallelizable operations (no matrix inversions), so the GPU is able to run the algorithm in constant time as the number of parameters in the model grows by a factor of 100.\vspace{0pt}}
    \label{fig:cpu_vs_gpu_heatmap}
\end{figure}\ignorespaces
\subsection{Parallelizing a Kalman Filter over Sensor Dimensions}
We now convert our decoupled optimization problems from \eqref{mini_opt} into a form such that a Kalman Filter can be used to solve for the estimates of each of the $n_y$ rows of $G$ in parallel. From here, a common structure amongst these $n_y$ filters allows us to handle the estimated $G$ directly in a single, parallelizable LML algorithm in which the most expensive operations are matrix multiplications.

To start, the estimate of each row of $G$ has its own Gaussian belief. Each of these $n_y$ beliefs are represented with a mean estimate $\hat{g}^{(j)}$ and covariance $\Sigma^{(j)}$. From here,
the discrete-time dynamics for the filter are:
\begin{align}
    g^{(j)}_{k+1} &= g^{(j)}_k + q^{(j)}_k, & q^{(j)} &\sim \mathcal{N}(0, Q),\label{lml_dynamics}
\end{align}
where $q^{(j)} \in \R{n_w}$ is the (unknown) additive Gaussian process noise, with a covariance of $Q \in \mathbf{S}_+^{n_w}$.  This can also be interpreted as a more expressive version of a ``forgetting factor'', where a non-zero process-noise covariance quantifies the expected change in parameters over the course of the experiment, therefore putting more emphasis on newer measurements than old. 

The measurement function in the filter for row $j$ is where we use the row parameters $g^{(j)}$ with our feature vector $w$ to predict the $j^{\text{th}}$ measurement:
\begin{align}
    y^{(j)}_k &=  w_k^T g^{(j)}_k + v_k, & v &\sim \mathcal{N}(0, 1.0), \label{lml_measurement}
\end{align}
where $v \in \R{}$ is the (unknown) sensor noise with a variance of one, a result that comes from our change of variables in \eqref{full_opt_ls_2}. Importantly, the measurement in \eqref{lml_measurement} is a scalar, which means the computation of the Kalman Gain only requires the inversion of a scalar instead of a matrix:
\begin{align}
    \ell = (\Sigma w)(w^T \Sigma w + 1)^{-1} = \frac{\Sigma w}{w^T \Sigma w + 1},
\end{align}
eliminating the only matrix inversion present in the Kalman Filter. 
\begin{figure}[t]
\begin{minipage}[t]{.54\textwidth}
 \SetAlCapHSkip{0em}
        \setlength{\algomargin}{0em}
        \begin{algorithm2e}[H]
        \caption{Linear Model Learning}
        \label{alg:lml}
        \DontPrintSemicolon
        \SetAlgoLined
        \SetNoFillComment
        \SetKwComment{Comment}{$\triangleright$\ }{} 
        \KwIn{$\hat{G}_{k|k}, \Sigma_{k|k}, w_{k+1}, y_{k+1}$}
        \tcc{predict belief updates}
        $\hat{G}_{k+1|k} \leftarrow \hat{G}_{k|k}$  \;
        $\Sigma_{k+1|k}\leftarrow \Sigma_{k|k} + Q$ \;
         \tcc{innovation and Kalman Gain}
        $z \leftarrow y_{k+1} - \hat{G}_{k+1|k} w_{k+1}$  \; 
        $s \leftarrow w_{k+1}^T \Sigma_{k+1|k} w_{k+1} + 1$ \; 
        $\ell \leftarrow \Sigma_{k+1|i} w_{k+1} / s $ \; 
        \tcc{\texttt{update mean and covariance}}
        $\hat{G}_{k+1|k+1} \leftarrow \hat{G}_{k+1|k} + z\ell^T  $ \;
        $\Sigma_{k+1|k+1} \leftarrow (I - \ell w_{k+1}^T)\Sigma_{k+1|k}(I - \ell w_{k + 1}^T) + \ell \ell^T$\;
        \KwOut{$\hat{G}_{k+1|k+1}, \Sigma_{k+1|k+1}$}
        \end{algorithm2e}
\end{minipage}%
\hspace{.3cm}
\begin{minipage}[t]{.43\textwidth}
\SetAlCapHSkip{0em}
        \setlength{\algomargin}{0em}
        \begin{algorithm2e}[H]
        \caption{Adaptive Regularization}
        \label{alg:adapt}
        \DontPrintSemicolon
        \SetAlgoLined
        \SetNoFillComment
        \SetKwComment{Comment}{$\triangleright$\ }{} 
        \KwIn{$\hat{G}, \Sigma, \rho$}
        \For{$i\leftarrow 1$ \KwTo $n_w$}{
        $c \leftarrow 0_{n_w}$ \; 
        $c[i] \leftarrow 1 $\;
        $r \leftarrow {(1}/{\rho[i])^2}$ \;
        $s \leftarrow \Sigma[i, i] + r$ \;
        $\ell \leftarrow \Sigma[:,i] / s$  \;
        $\Sigma \leftarrow (I - \ell c^T)\Sigma(I - \ell c^T) + r\ell \ell^T$\;
        $\hat{G}\leftarrow \hat{G} -\hat{G}[:,i]\ell^T$
        }
        \KwOut{$\hat{G}, \Sigma$}
        \end{algorithm2e}
\end{minipage}
\end{figure}
\noindent

The dynamics and measurement equations in (\ref{lml_dynamics}-\ref{lml_measurement}) are in the canonical form shown in (\ref{vanilla_kf_1}-\ref{vanilla_kf_2}), where $A = I$ and $C = w_k^T$.  Critically missing from these matrices is any reference to row $j$, meaning the filters for all $n_y$ rows have the same process and measurement functions. This enables two important commonalities amongst the $n_y$ filters: They all share the same covariance and Kalman Gain. Therefore, we only need to maintain a single ``global'' covariance $\Sigma$ that represents each of the $n_y$ rows, and use the shared Kalman Gain, $\ell \in \R{n_y}$ to update everything.

Using the Kalman Gain, the updates to the mean estimate of each row are:
\begin{align}
    \hat{g}^{(j)}_{k+1|k+1} = \hat{g}^{(j)}_{k+1|k} + \ell (y_{k+1}^{(j)} - \hat{g}^{(j)}_{k+1|k} w_{k+1}) \quad \quad \quad \forall j \in [1, \,\,n_y],
\end{align}
and since the Kalman Gain is shared, these individual row updates can be broadcast to an estimate of the full matrix $\hat{G}$ :
\begin{align}
    \hat{G}_{k+1|k+1} = \hat{G}_{k+1|k} + (y_{k+1} - \hat{G}_{k+1|k} w_{k+1})\ell^T,
\end{align}
where the parallelism is simply handled by a vector outer product and matrix addition. Using this technique on the parallel filters results in the LML algorithm, which is detailed in Algorithm \ref{alg:lml}.

Lastly, the regularization term from \eqref{mini_opt} must be included in the LML algorithm as an initialization. To do this, the cost terms associated with the initial Gaussian belief in the MAP problem \eqref{map} are adapted to replicate the regularization in \eqref{mini_opt} in the form of an initial Gaussian belief over row $j$:
\begin{align}
    \hat{g}^{(j)}_{0|0} &= 0_{n_w}, & \Sigma_{0|0} &= \operatorname{diag}(1 / b^2). \label{eq:init_reg}
\end{align}
With this initialization, the LML algorithm is called recursively every time a new measurement is available. The entire algorithm is matrix-inversion free, requiring only simple matrix products and additions that are trivial to parallelize on a GPU. To demonstrate the importance of this, timing results of the LML algorithm run on a GPU and CPU are shown in Fig. \ref{fig:cpu_vs_gpu_heatmap}, where the run time on a GPU stays constant as the number of parameters in $\hat{G}$ grows by a factor of 100. 
\begin{SCfigure}[][t]
    \centering
    \input{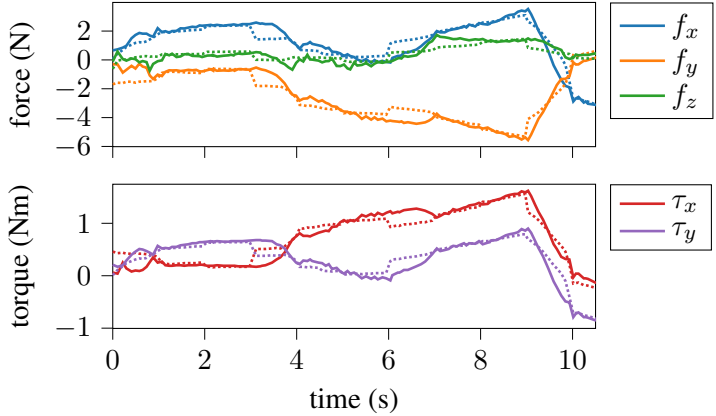}
    \caption{The sim-to-real gap for a connector insertion calibration sequence, where the real force torque sensor measurements are solid and the predicted measurements from our learned model are dotted.  The learned model is able to accurately capture the contact force behavior of the connector inside the socket.}
    \label{fig:sim2real}
    \vspace{0pt}
\end{SCfigure}
\noindent
\subsection{Adaptive Regularization}
In the original cost function of \eqref{full_opt}, the relative importance of the regularization term is dependent on the trajectory length.  As the trajectory gets longer, the impact the regularization has on the estimate decreases. In cases where the LML algorithm is being run for an unknown amount of time, choosing the appropriate regularizer offline is difficult. To deal with this, we present a method for updating the regularization term online that can handle arbitrary quadratic regularization and does not require any matrix inversions. 

We specify this additional regularizer in a general form $\|G\rho\|_2^2$ that will be included in the cost function of \eqref{full_opt}. To add this to the LML algorithm, the connection between the MAP problem in \eqref{map} and the Kalman Filter is used once again. As before, we formulate a fictitious measurement equation for each $\hat{g}^{(j)}$ such that we recover the quadratic regularizer in the MAP cost, then use the same parallelization strategy to generalize it to $\hat{G}$.

Our estimated mean $\hat{g}^{(j)}$ will have stationary dynamics, but we use a measurement function where each row has the values of the row $j$ returned,
\begin{align}
    z_{k} &= \hat{g}^{(j)}_k + r_k\label{kf_reg_2}.
\end{align}
By expecting this ``measurement'' to be zero, we use our regularizer as a covariance for the fictitious sensor, $r\sim \mathcal{N}(0, \operatorname{diag}(1/\rho^2))$, reconstructing our desired regularization cost in the MAP problem.  Normally the Kalman Filter requires a matrix the size of the measurement to be inverted, however, since this fictitious sensor noise is diagonal, we can use the method of sequential scalar measurement updates to once again avoid any matrix inversions \citep{kailath2000}. Similar to the LML algorithm, these operations can be broadcast to each row of $G$ using matrix multiplication, as shown in Algorithm \ref{alg:adapt}.
%
\section{Experiments}
%
We demonstrate LML algorithm on a contact-rich connector-insertion task, where the resulting contact-force model is used in a convex-optimization control policy. This approach is first shown to work in simulation with MuJoCo, then on hardware with a Franka robot arm with a parallel gripper holding a standard C13 power plug.

To learn the relationship between the configuration, control, and contact wrench on the plug, the LML algorithm will be used with the following feature vector:
\begin{align}
    w = \operatorname{vcat}(r,\, \operatorname{vec}(R),\, r_{des},\, \phi,\, 1),
\end{align}
where $r\in \R{3}$ is the cartesian position of the end effector, $R \in \mathbf{SO(3)}$ is the rotation matrix describing the attitude of the end effector, $r_{des} \in \R{3}$ is the desired position sent to an impedance controller, $\phi \in \R{3}$ is the axis-angle rotation between current attitude and desired attitude, and the $1$ is included to account for a constant bias. The control inputs in this feature vector are $r_{des}$ and $\phi$, where the axis-angle parameterization is used to avoid any attitude-related constraints \citep{jackson2021a}. 
%
\noindent
\begin{figure}[t]
\centering
\begin{minipage}{.3\textwidth}
  \centering
  \includegraphics[width=\linewidth]{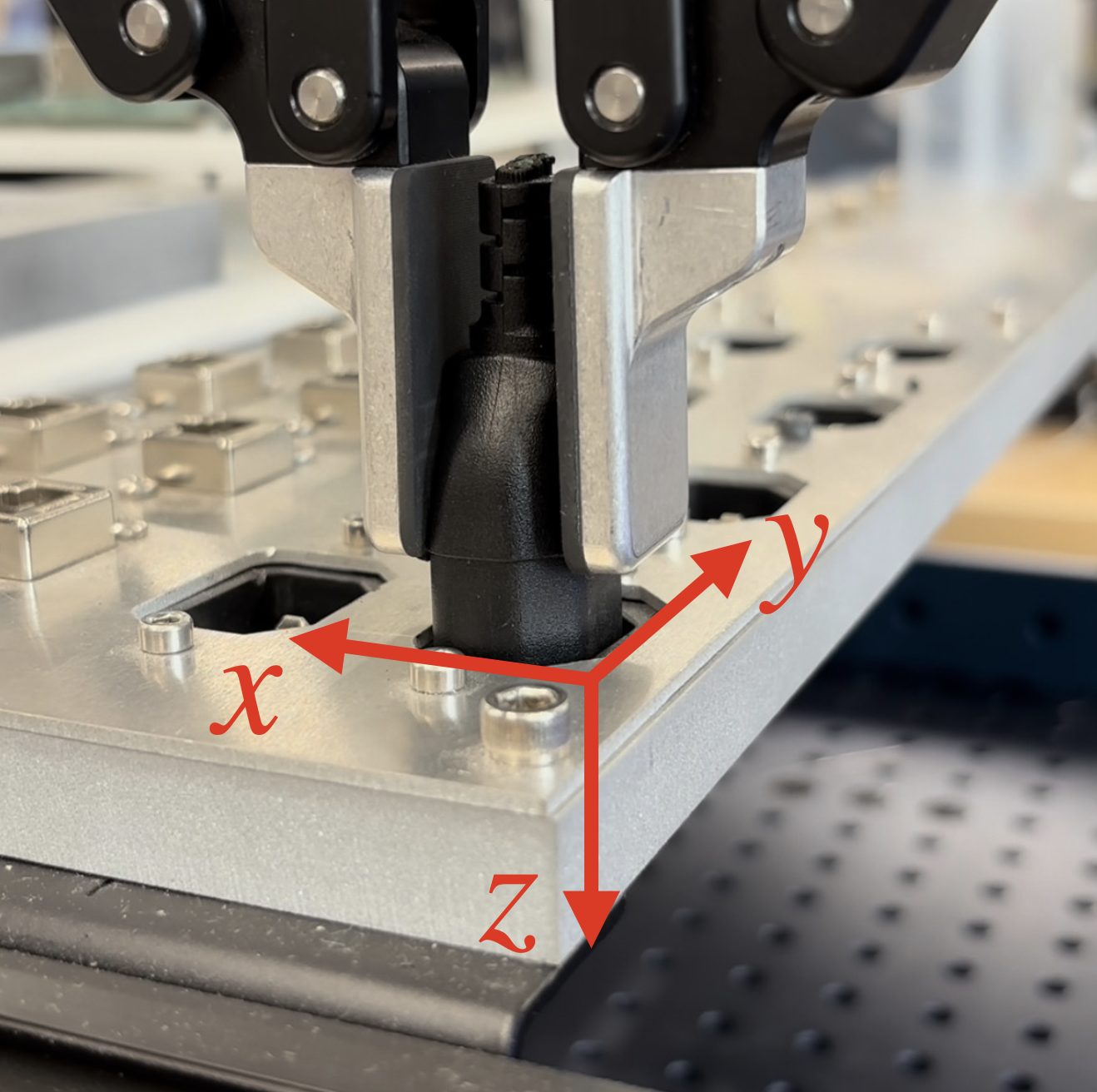}
  \captionof{figure}{Calibration and insertion of a power plug into a socket with a Franka robot arm and parallel gripper.}
  \label{fig:plug}
\end{minipage}
\hspace{.8cm}
\begin{minipage}{.6\textwidth}
  \centering
  \input{figures/force_during_experiment.tex}
  \vspace{-10pt}
    \captionof{figure}{Magnitude of the force on the plug in the XY axis during insertion of eight different plugs. The scripted insertion leaves a constant force on the plug, after which the policy is used to reduce the force on the connector.}
  \label{fig:force_during_experiment}
\end{minipage}%
\end{figure}

The LML algorithm learns a linear relationship between this feature vector and the force torque sensor, $y \approx G w$. To do this, a calibration sequence is run where both the position and attitude are varied. The force-torque sensor measurements and feature vectors are input to the LML algorithm during this calibration, and the resulting estimated sensor measurements are able to closely align with the true sensor measurements as shown in Fig. \ref{fig:sim2real}.
To align the connector for insertion, a model-based controller uses this linear relationship to solve for controls that minimize the resistance measured by the force-torque sensor. The controller is posed as the solution to the following optimization problem:
\begin{equation}
    \begin{array}{ll}
    \underset{r_{des},\, \phi}{\mbox{minimize}} & \|S\hat{y}\|_2^2 + \lambda \|r - r_{des}\|_2^2 + \mu \| \phi \|_2^2\\ 
    \mbox{subject to} &  \hat{y} = G [r^T,\, \operatorname{vec}(R)^T,\, r_{des}^T,\, \phi^T,\, 1]^T,\\ 
    & \|r - r_{des}\|_\infty \leq 1 \, \, \text{cm}, \\ 
    & \|\phi\|_\infty \leq 3^\circ ,
    \end{array} \label{mdp2}
\end{equation}
where $S = \operatorname{diag}([1,1,0,1,1,1])$ is used to penalize all forces and torques except for force in the Z (plug insertion) direction, $\lambda \in \R{+}$ and $\mu \in \R{+}$ are regularization weights, and the primal constraints ensure the commanded end-effector pose doesn't deviate too far from the current pose. Due to the linearity of our model, this problem is a small convex quadratic program, for which there are many available solvers for fast online execution \citep{mattingley2012}. We have found that a solver is not necessary since the controls can be sub-optimally clipped in the rare case the constraints are violated.

This system-identification and control strategy was first demonstrated in MuJoCo, as seen in Fig. \ref{fig:mujoco}, where the controller was successfully able to align the plug with initial misalignments. Second, a Franka robot arm with a parallel gripper was able to insert the C13 power plug into eight different sockets in the presence of misalignment, sensor noise, and actuator uncertainty. In both of these scenarios, LML was run on the force-torque sensor data during a quick 10-second calibration sequence, after which the controller was turned on and smooth insertion was achieved. On the Franka, the forces on the plug during these eight alignments are shown in Fig. \ref{fig:force_during_experiment}, where the forces from the initial misalignment are dramatically reduced by over 80\% once the controller takes over.
%
\section{Conclusion}
%
This work proposes a new approach for the modeling and control of connector-insertion tasks, where an estimated quasi-static contact model is used to predict the contact forces on the connector as a function of the configuration and control. This contact force model is learned online to global optimality with a novel Linear Model Learning algorithm that is free of matrix inversions. The runtime of this algorithm is shown to be constant on a GPU as the model size increases, since it is entirely comprised of parallelizable operations. The effectiveness of this modeling strategy is demonstrated both in simulation and in hardware experiments, where the contact-force model is used to solve for an optimal alignment policy for C13 power connectors. 
\newpage 
%
%
\bibliography{rexlab.bib}

\begin{thebibliography}{37}
\providecommand{\natexlab}[1]{#1}
\providecommand{\url}[1]{\texttt{#1}}
\expandafter\ifx\csname urlstyle\endcsname\relax
  \providecommand{\doi}[1]{doi: #1}\else
  \providecommand{\doi}{doi: \begingroup \urlstyle{rm}\Url}\fi

\bibitem[Acary et~al.(2018)Acary, Br{\'e}mond, and Huber]{acary2018}
Vincent Acary, Maurice Br{\'e}mond, and Olivier Huber.
\newblock On {{Solving Contact Problems}} with {{Coulomb Friction}}:
  {{Formulations}} and {{Numerical Comparisons}}.
\newblock In Remco Leine, Vincent Acary, and Olivier Br{\"u}ls, editors,
  \emph{Advanced {{Topics}} in {{Nonsmooth Dynamics}}}, pages 375--457.
  {Springer International Publishing}, {Cham}, 2018.
\newblock ISBN 978-3-319-75971-5 978-3-319-75972-2.
\newblock \doi{10.1007/978-3-319-75972-2_10}.

\bibitem[Benesty and Gansler(2005)]{benesty2005}
J.~Benesty and T.~Gansler.
\newblock A recursive estimation of the condition number in the {{RLS}}
  algorithm [adaptive signal processing applications].
\newblock In \emph{Proceedings. ({{ICASSP}} '05). {{IEEE International
  Conference}} on {{Acoustics}}, {{Speech}}, and {{Signal Processing}}, 2005.},
  volume~4, pages iv/25--iv/28 Vol. 4, March 2005.
\newblock \doi{10.1109/ICASSP.2005.1415936}.

\bibitem[Boyd and Vandenberghe(2004)]{boyd2004}
Stephen Boyd and Lieven Vandenberghe.
\newblock \emph{Convex {{Optimization}}}.
\newblock {Cambridge University Press}, 2004.

\bibitem[Chatterjee()]{chatterjee}
Anindya Chatterjee.
\newblock On the {{Realism}} of {{Complementarity Conditions}} in {{Rigid Body
  Collisions}}.
\newblock page~10.

\bibitem[Coumans(2015)]{coumans2015}
Erwin Coumans.
\newblock Bullet {{Physics Simulation}}.
\newblock In \emph{{{SIGGRAPH}}}, {Los Angeles}, 2015. {ACM}.
\newblock ISBN 978-1-4503-3634-5.
\newblock \doi{10.1145/2776880.2792704}.

\bibitem[Drumwright et~al.(2010)Drumwright, Shell, Drumwright, and
  Shell]{drumwright2010}
Evan Drumwright, Dylan~A. Shell, Evan Drumwright, and Dylan~A. Shell.
\newblock Modeling {{Contact Friction}} and {{Joint Friction}} in {{Dynamic
  Robotic Simulation Using}} the {{Principle}} of {{Maximum Dissipation}}.
\newblock In Frans Groen, David Hsu, Volkan Isler, Jean-Claude Latombe, and
  Ming~C. Lin, editors, \emph{Wokrshop on the {{Algorithmic Foundations}} of
  {{Robotics}} ({{WAFR}})}, {Singapore}, 2010.
\newblock ISBN 978-3-642-17451-3 978-3-642-17452-0.
\newblock \doi{10.1007/978-3-642-17452-0_15}.

\bibitem[Elandt et~al.(2019)Elandt, Drumwright, Sherman, and Ruina]{elandt2019}
Ryan Elandt, Evan Drumwright, Michael Sherman, and Andy Ruina.
\newblock A pressure field model for fast, robust approximation of net contact
  force and moment between nominally rigid objects.
\newblock \emph{arXiv:1904.11433 [cs]}, April 2019.

\bibitem[Fu et~al.(2023)Fu, Huang, Berscheid, Li, Goldberg, and Chitta]{fu2023}
Letian Fu, Huang Huang, Lars Berscheid, Hui Li, Ken Goldberg, and Sachin
  Chitta.
\newblock Safe {{Self-Supervised Learning}} in {{Real}} of {{Visuo-Tactile
  Feedback Policies}} for {{Industrial Insertion}}, March 2023.

\bibitem[Galrinho(2016)]{galrinho2016}
Miguel Galrinho.
\newblock Least {{Squares Methods}} for {{System Identification}} of
  {{Structured Models}}.
\newblock 2016.

\bibitem[Ghahramani()]{ghahramani}
Zoubin Ghahramani.
\newblock Parameter {{Estimation}} for {{Linear Dynamical Systems}}.

\bibitem[Halm and Posa(2019)]{halm2019}
Mathew Halm and Michael Posa.
\newblock A {{Quasi-static Model}} and {{Simulation Approach}} for {{Pushing}},
  {{Grasping}}, and {{Jamming}}, February 2019.

\bibitem[Hardt et~al.(2019)Hardt, Ma, and Recht]{hardt2019}
Moritz Hardt, Tengyu Ma, and Benjamin Recht.
\newblock Gradient {{Descent Learns Linear Dynamical Systems}}, February 2019.

\bibitem[Hazan et~al.(2017)Hazan, Singh, and Zhang]{hazan2017}
Elad Hazan, Karan Singh, and Cyril Zhang.
\newblock Learning {{Linear Dynamical Systems}} via {{Spectral Filtering}},
  November 2017.

\bibitem[Ho and Kalman(1966)]{ho1966}
{$B$}~L. Ho and R.~E. Kalman.
\newblock {Editorial: Effective construction of linear state-variable models
  from input/output functions: Die Konstruktion von linearen Modeilen in der
  Darstellung durch Zustandsvariable aus den Beziehungen f\"ur Ein- und
  Ausgangsgr\"o\ss en}.
\newblock \emph{at - Automatisierungstechnik}, 14\penalty0 (1-12):\penalty0
  545--548, December 1966.
\newblock ISSN 2196-677X.
\newblock \doi{10.1524/auto.1966.14.112.545}.

\bibitem[Hoburg and Tedrake(2009)]{hoburg2009}
Warren Hoburg and Russ Tedrake.
\newblock System {{Identification}} of {{Post Stall Aerodynamics}} for {{UAV
  Perching}}.
\newblock In \emph{{{AIAA Infotech}}@{{Aerospace Conference}}}. {American
  Institute of Aeronautics and Astronautics}, 2009.

\bibitem[Horak and Trinkle(2019)]{horak2019}
Peter~C. Horak and Jeff~C. Trinkle.
\newblock On the {{Similarities}} and {{Differences Among Contact Models}} in
  {{Robot Simulation}}.
\newblock \emph{IEEE Robotics and Automation Letters}, 4\penalty0 (2):\penalty0
  493--499, April 2019.
\newblock ISSN 2377-3766, 2377-3774.
\newblock \doi{10.1109/LRA.2019.2891085}.

\bibitem[Howell et~al.(2022)Howell, Cleac'h, Kolter, Schwager, and
  Manchester]{howell2022}
Taylor~A. Howell, Simon~Le Cleac'h, J.~Zico Kolter, Mac Schwager, and Zachary
  Manchester.
\newblock Dojo: {{A Differentiable Physics Engine}} for {{Robotics}}.
\newblock \emph{IEEE Transactions on Robotics and Automation (In Review)}, June
  2022.
\newblock \doi{10.48550/arXiv.2203.00806}.

\bibitem[Jackson et~al.(2021)Jackson, Tracy, and Manchester]{jackson2021a}
Brian~E Jackson, Kevin Tracy, and Zachary Manchester.
\newblock Planning with {{Attitude}}.
\newblock \emph{IEEE Robotics and Automation Letters}, January 2021.

\bibitem[Kailath et~al.(2000)Kailath, Sayed, and Hassibi]{kailath2000}
Thomas Kailath, Ali~H. Sayed, and Babak Hassibi.
\newblock \emph{Linear {{Estimation}}}.
\newblock {Prentice Hall}, 2000.
\newblock ISBN 978-0-13-022464-4.

\bibitem[Kalman(1960)]{kalman1960}
R.~E. Kalman.
\newblock A {{New Approach}} to {{Linear Filtering}} and {{Prediction
  Problems}}.
\newblock \emph{Journal of Basic Engineering}, 82\penalty0 (1):\penalty0
  35--45, March 1960.
\newblock ISSN 0021-9223.
\newblock \doi{10.1115/1.3662552}.

\bibitem[Kang et~al.(2022)Kang, Zang, Wang, and Chen]{kang2022}
Hanwen Kang, Yaohua Zang, Xing Wang, and Yaohui Chen.
\newblock Uncertainty-{{Driven Spiral Trajectory}} for {{Robotic Peg-in-Hole
  Assembly}}.
\newblock \emph{IEEE Robotics and Automation Letters}, 7\penalty0 (3):\penalty0
  6661--6668, July 2022.
\newblock ISSN 2377-3766.
\newblock \doi{10.1109/LRA.2022.3176718}.

\bibitem[Lee et~al.(2018)Lee, X.~Grey, Ha, Kunz, Jain, Ye, S.~Srinivasa,
  Stilman, and Karen~Liu]{lee2018}
Jeongseok Lee, Michael X.~Grey, Sehoon Ha, Tobias Kunz, Sumit Jain, Yuting Ye,
  Siddhartha S.~Srinivasa, Mike Stilman, and C.~Karen~Liu.
\newblock {{DART}}: {{Dynamic Animation}} and {{Robotics Toolkit}}.
\newblock \emph{The Journal of Open Source Software}, 3\penalty0 (22):\penalty0
  500, February 2018.
\newblock ISSN 2475-9066.
\newblock \doi{10.21105/joss.00500}.

\bibitem[Liavas and Regalia(1998)]{liavas1998}
A.P. Liavas and P.A. Regalia.
\newblock Numerical stability issues of the conventional recursive least
  squares algorithm.
\newblock In \emph{Proceedings of the 1998 {{IEEE International Conference}} on
  {{Acoustics}}, {{Speech}} and {{Signal Processing}}, {{ICASSP}} '98 ({{Cat}}.
  {{No}}.{{98CH36181}})}, volume~3, pages 1409--1412 vol.3, May 1998.
\newblock \doi{10.1109/ICASSP.1998.681711}.

\bibitem[Liavas and Regalia(1999)]{liavas1999}
A.P. Liavas and P.A. Regalia.
\newblock On the numerical stability and accuracy of the conventional recursive
  least squares algorithm.
\newblock \emph{IEEE Transactions on Signal Processing}, 47\penalty0
  (1):\penalty0 88--96, January 1999.
\newblock ISSN 1941-0476.
\newblock \doi{10.1109/78.738242}.

\bibitem[Makoviychuk et~al.(2021)Makoviychuk, Wawrzyniak, Guo, Lu, Storey,
  Macklin, Hoeller, Rudin, Allshire, Handa, and State]{makoviychuk2021}
Viktor Makoviychuk, Lukasz Wawrzyniak, Yunrong Guo, Michelle Lu, Kier Storey,
  Miles Macklin, David Hoeller, Nikita Rudin, Arthur Allshire, Ankur Handa, and
  Gavriel State.
\newblock Isaac {{Gym}}: {{High Performance GPU-Based Physics Simulation For
  Robot Learning}}, August 2021.

\bibitem[Mason(2001)]{mason2001a}
Matthew~T. Mason.
\newblock \emph{Mechanics of {{Robotic Manipulation}}}.
\newblock {The MIT Press}, June 2001.
\newblock ISBN 978-0-262-25662-9.
\newblock \doi{10.7551/mitpress/4527.001.0001}.

\bibitem[Mattingley and Boyd(2012)]{mattingley2012}
Jacob Mattingley and Stephen Boyd.
\newblock {{CVXGEN}}: A code generator for embedded convex optimization.
\newblock In \emph{Optimization {{Engineering}}}, pages 1--27, 2012.

\bibitem[Nair et~al.(2023)Nair, Zhu, Narayanan, Solowjow, and Levine]{nair2023}
Ashvin Nair, Brian Zhu, Gokul Narayanan, Eugen Solowjow, and Sergey Levine.
\newblock Learning on the {{Job}}: {{Self-Rewarding Offline-to-Online
  Finetuning}} for {{Industrial Insertion}} of {{Novel Connectors}} from
  {{Vision}}, February 2023.

\bibitem[Pang and Tedrake(2018)]{pang2018}
Tao Pang and Russ Tedrake.
\newblock A {{Robust Time-Stepping Scheme}} for {{Quasistatic Rigid Multibody
  Systems}}.
\newblock In \emph{2018 {{IEEE}}/{{RSJ International Conference}} on
  {{Intelligent Robots}} and {{Systems}} ({{IROS}})}, pages 5640--5647,
  {Madrid}, October 2018. {IEEE}.
\newblock ISBN 978-1-5386-8094-0.
\newblock \doi{10.1109/IROS.2018.8594378}.

\bibitem[Pang et~al.(2023)Pang, Suh, Yang, and Tedrake]{pang2023}
Tao Pang, H.~J.~Terry Suh, Lujie Yang, and Russ Tedrake.
\newblock Global {{Planning}} for {{Contact-Rich Manipulation}} via {{Local
  Smoothing}} of {{Quasi-dynamic Contact Models}}, February 2023.

\bibitem[Suh and Tedrake(2020)]{suh2020}
H.~J.Terry Suh and Russ Tedrake.
\newblock The {{Surprising Effectiveness}} of {{Linear Models}} for {{Visual
  Foresight}} in {{Object Pile Manipulation}}.
\newblock \emph{Springer Proceedings in Advanced Robotics}, 17:\penalty0
  347--363, February 2020.
\newblock \doi{10.48550/arxiv.2002.09093}.

\bibitem[Tang et~al.(2023)Tang, Lin, Akinola, Handa, Sukhatme, Ramos, Fox, and
  Narang]{tang2023}
Bingjie Tang, Michael~A. Lin, Iretiayo Akinola, Ankur Handa, Gaurav~S.
  Sukhatme, Fabio Ramos, Dieter Fox, and Yashraj Narang.
\newblock {{IndustReal}}: {{Transferring Contact-Rich Assembly Tasks}} from
  {{Simulation}} to {{Reality}}, May 2023.

\bibitem[Tedrake(2014)]{tedrake2014}
Russ Tedrake.
\newblock Underactuated {{Robotics}}: {{Algorithms}} for {{Walking}},
  {{Running}}, {{Swimming}}, {{Flying}}, and {{Manipulation}} ({{Course Notes}}
  for {{MIT}} 6.832).
\newblock Technical report, {Downloaded in Fall, 2014 from
  http://people.csail.mit.edu/russt/underactuated/}, 2014.

\bibitem[Todorov et~al.(2012)Todorov, Erez, and Tassa]{todorov2012}
E.~Todorov, T.~Erez, and Y.~Tassa.
\newblock {{MuJoCo}}: {{A}} physics engine for model-based control.
\newblock In \emph{2012 {{IEEE}}/{{RSJ International Conference}} on
  {{Intelligent Robots}} and {{Systems}}}, pages 5026--5033, October 2012.
\newblock \doi{10.1109/IROS.2012.6386109}.

\bibitem[Xinjilefu et~al.(2014)Xinjilefu, Feng, and Atkeson]{xinjilefu2014}
X~Xinjilefu, Siyuan Feng, and Christopher~G. Atkeson.
\newblock Dynamic state estimation using {{Quadratic Programming}}.
\newblock In \emph{2014 {{IEEE}}/{{RSJ International Conference}} on
  {{Intelligent Robots}} and {{Systems}}}, pages 989--994, {Chicago, IL, USA},
  September 2014. {IEEE}.
\newblock ISBN 978-1-4799-6934-0 978-1-4799-6931-9.
\newblock \doi{10.1109/IROS.2014.6942679}.

\bibitem[Zhao et~al.(2022)Zhao, Luo, Sushkov, Pevceviciute, Heess, Scholz,
  Schaal, and Levine]{zhao2022}
Tony~Z. Zhao, Jianlan Luo, Oleg Sushkov, Rugile Pevceviciute, Nicolas Heess,
  Jon Scholz, Stefan Schaal, and Sergey Levine.
\newblock Offline {{Meta-Reinforcement Learning}} for {{Industrial Insertion}},
  September 2022.

\bibitem[Zhao et~al.(2020)Zhao, Queralta, and Westerlund]{zhao2020}
Wenshuai Zhao, Jorge~Pe{\~n}a Queralta, and Tomi Westerlund.
\newblock Sim-to-{{Real Transfer}} in {{Deep Reinforcement Learning}} for
  {{Robotics}}: A {{Survey}}.
\newblock In \emph{2020 {{IEEE Symposium Series}} on {{Computational
  Intelligence}} ({{SSCI}})}, pages 737--744, December 2020.
\newblock \doi{10.1109/SSCI47803.2020.9308468}.

\end{thebibliography}
\end{document}